# SSP-Based Construction of Evaluation-Annotated Data for Fine-Grained Aspect-Based Sentiment Analysis

**Suwon Choi**
DICORA, Hankuk University of Foreign Studies
soown0607@gmail.com

**Shinwoo Kim**
DICORA, Hankuk University of Foreign Studies
siinwoo0306@gmail.com

**Changhoe Hwang**
DICORA, Hankuk University of Foreign Studies
hch8357@naver.com

**Gwanghoon Yoo**
DICORA, Hankuk University of Foreign Studies
rhkdgns2008@naver.com

**Eric Laporte**
LIGM, Université Gustave Eiffel
eric.laporte@univ-eiffel.fr

**Jeesun Nam**
DICORA, Hankuk University of Foreign Studies
namjs@hufs.ac.kr

## Abstract

We report the construction of a Korean evaluation-annotated corpus, hereafter called 'Evaluation Annotated Dataset (EVAD)', and its use in Aspect-Based Sentiment Analysis (ABSA) extended in order to cover e-commerce reviews containing sentiment and non-sentiment linguistic patterns. The annotation process uses Semi-Automatic Symbolic Propagation (SSP). We built extensive linguistic resources formalized as a Finite-State Transducer (FST) to annotate corpora with detailed ABSA components in the fashion e-commerce domain. The ABSA approach is extended, in order to analyze user opinions more accurately and extract more detailed features of targets, by including aspect values in addition to topics and aspects, and by classifying aspect-value pairs depending whether values are unary, binary, or multiple. For evaluation, the KoBERT and KcBERT models are trained on the annotated dataset, showing robust performances of F1 0.88 and F1 0.90, respectively, on recognition of aspect-value pairs.

## 1 Introduction

Aspect-Based Sentiment Analysis (ABSA) is a sentiment analysis technique which extracts users' opinions on the basis of opinion quintuples consisting of an entity or target, an aspect, a sentiment value, an opinion holder, and a time (Liu 2012, 2015). Since the opinion holder and time can be identified by meta-information sources such as user ids and content posting time, opinion triples including an entity (e), an aspect (a), and a sentiment value (s) are the main components to be extracted from texts associating a semantic orientation with aspects of a target product or service. For example, in (1), the opinion triple plays a vital role in identifying a user's 'positive' (s) sentiment attributed to the 'design' (a) aspect of a 'jacket' (e) target.

(1) [이 자켓]은 [디자인]이 [괜찮아요]
*[i cakheys]un [ticain]i [kwaynchanhayo]*
The **design** of **this jacket** is **suitable** for me.
(e:jacket, a:design, s:positive)

However, current ABSA systems face limitations in processing texts with Multi-Word Expressions (MWE). For example, (2) shows a positive opinion about a target 자켓 (*cakheys*) 'jacket' with an MWE 마음에 들어요 (*maumey tuleyo*) 'fit the bill; suitable', literally 'listen in the heart', though none of these words conveys a positive value: the MWE is non-compositional, i.e. it functions as a single word.

(2) [이 자켓]은 [긴] [길이]가 [마음에 들어요]
*[i cakheys]un [kin] [kili]ka [maumey tuleyo]*
The **longer length** of **this jacket** **fits the bill**.

(e:jacket, a:length-long, s:positive)

Additionally, traditional ABSA opinion triples fail to handle detailed information related to aspects, especially when it involves non-polar opinion expressions. For instance, the aspect 길이 (*kili*) 'length' has a specific value 길다 (*kilta*) 'long', which has pivotal relevance to fine-grained ABSA, as it explains why the aspect has the polarity expressed in the sentence, but this value qualifies neither as aspect nor as sentiment, since it has no sentiment polarity in itself.

In order to cover detailed information, this study uses aspect-value pairs as an enhancement to ABSA, and describes the Evaluation Annotated Dataset (EVAD), which systematically classifies and formalizes detailed opinion elements in online e-commerce domains. We adapt traditional ABSA opinion triples by adding a level of analysis defined by 'evaluation triples' (ET) (Nam, 2021a), which consist of three components: **topic**, **aspect**, and **value**. The **topic** and **aspect** are the target and aspect of traditional e/a/s opinion triples. The **value**, however, is a new element for capturing the value of aspects, no matter whether they are realized in the form of a word or of an MWE.

Either the aspect and value appear together in a complex phrase (i.e. aspect-value pair), as shown in example (2), or the aspect term is simply omitted, so we suggest to consider each aspect-value pair as a single item, and to view an aspect term or a value expression as a particular case of such an item as well. In this way, the value, which can be sentiment-oriented or not, may constitute a complex aspect phrase or a sentiment predicate. In this study, aspect-value pairs are categorized according to their semantic characteristics. The first type, Unary, includes existential predicates such as 있다 (*issa*) 'be present' and 없다 (*epsta*) 'be absent', and the polar orientation depends on which Unary aspect occurs. For instance, an aspect-value pair 방수성이 있다 (*pangswusengi issa*) 'waterproofness is present' expresses a positive polarity.

The second type, Binary, contains aspects with binary value, which may convey sentiment or non-sentiment information such as 길이가 길다 (*kilika kilta*) 'longer length' or 짧지 않다 (*ccalpci anhta*) 'not short'. Differently from the first type, the sentiment orientation is not determined by the aspect-value pair itself. When (2) is analyzed with an ET, an aspect-value pair 긴 길이 (*kin kili*) 'longer length', classified as Binary and undetermined in polarity, appears as a complex aspect. As it is associated with a positive predicate, it is subsequently analyzed as 'e:this jacket, a:length-long, s:length-positive.'

Finally, aspects with more values, such as 색깔이 빨강색이에요 (*saykkkali ppalkangsaykieyyo*) 'the color is red', are classified as Multiple. This type of the aspect-value, undetermined in sentiment orientation, may occur as a complex aspect phrase. When sentiment expressions collocate with it, the detailed information is extracted as an aspect-value pair and a sentiment value.

The main contribution of this study is to test a method to extract more intricate information than traditional ABSA, by annotating information about aspect and value simultaneously, without a separate pairing process. In order to implement the notion of evaluation triple presented above into the actual dataset, we built EVAD through the analysis of a massive online clothing review corpus. To build this resource, we used the Semi-Automatic Symbolic Propagation (SSP) method (Nam, 2021b; Hwang et al., 2021) with Local Grammar Graphs (LGGs) compiled into a Finite-State Transducer (Gross, 1997, 1999) and the Korean lexical databases 'Dictionnaire Électronique du Coréen' (DECO) and 'Deco-Dom' (Nam, 2018), to analyze corpora on the Unitex platform (Paumier, 2003). Figure 1 below presents the overall flow chart for the construction of EVAD.

## 2 Related studies

As several deep learning models have been used to perform ABSA, many researchers have developed datasets to train the models.

Jiang et al. (2019) points out that in existing benchmark data, most examples are not significantly different from sentence-level sentiment analysis because they contain only one aspect, or all aspects have the same sentiment polarity. Therefore, the study introduces MAMS, a dataset for more sophisticated ABSA, allowing all examples to contain two or more polarities.

The dataset is extracted from the same corpus as the existing benchmark data.

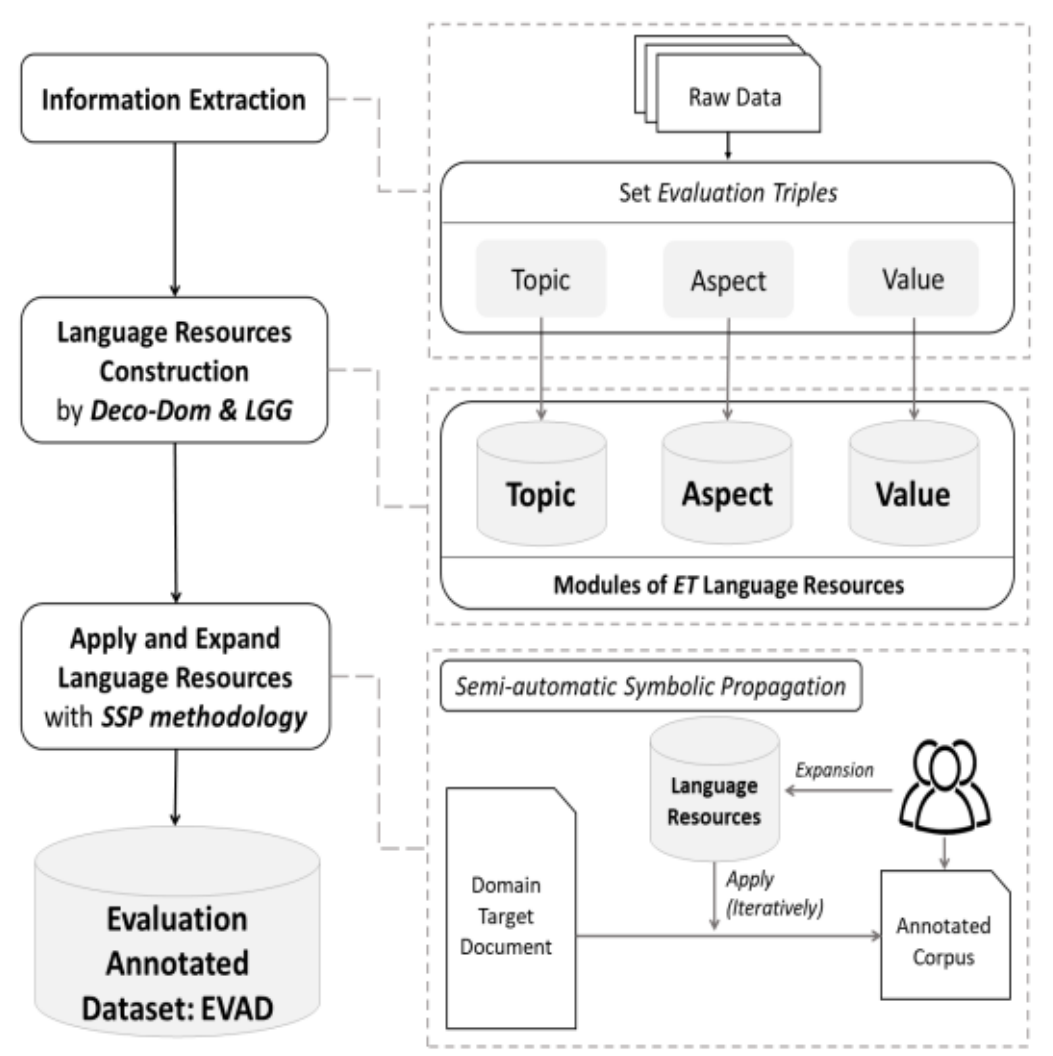

Figure 1: Flow chart for construction of EVAD

Orbach et al. (2020) notices that existing sentiment analysis studies use data from a limited range of domains, and presents YASO, a dataset from Yelp, Amazon, SST (Socher et al., 2013), and Opinosis (Ganesan et al., 2010) for ABSA learning and evaluation. The YASO dataset covers diverse domains and allows for measuring cross-domain performance. However, it does not provide meta-information on the domains of the reviews.

Other studies define new problems for ABSA. For example, Fan et al. (2019) points out that it is essential to find which words convey the sentiment related to a specific aspect. Therefore, a task called TOWE is defined and a dedicated dataset is provided. Peng et al. (2020) uses TOWE and SemEval data and argues that complete ABSA requires pairing aspects with the corresponding 'opinion terms', i.e. the 'why' expressions that give a clue of why the aspect has the polarity expressed in the sentence, e.g. *friendly* in *waiters/friendly*. These 'opinion terms' describe a value of the aspect.

In some state-of-the-art datasets, expressions denoting aspect values are annotated, but only when they also convey sentiment information by themselves, as in *waiters/friendly*. Our study differs from that practice in that it includes in its subject non-sentiment information about aspect values, when that information is relevant to sentiment analysis.

SSP is a method of generating learning datasets by annotating selected expressions in collected data such as reviews, discussions or questions in the relevant domain in social media. This technology aims to enhance learning data with the aid of sophisticated, large-scale language resources. Once the first dataset is semi-automatically annotated, the language resources are edited based on this first result and applied to the corpus. This bootstrapping approach (Gross, 1999) is applied to the first and second versions of the annotated dataset, producing a third version (Nam, 2021b).

## 3 Method with Evaluation Triples

In section 3, we propose a method of sentiment analysis that focuses on evaluation triples (ET) and that we have applied to clothing reviews. ETs allow for classifying purchasers' objective or subjective evaluations on aspects such as color, material, pattern..., no matter whether the evaluation is expressed with or without an explicit sentiment polarity. An ET comprises a topic, an aspect, and a value. In the clothing domain, a topic is an evaluation target such as 티셔츠 (*thisyechu*) 'T-shirts' or 바지 (*paci*) 'pants'. An aspect of an evaluation target can be 가격 (*kakyek*) 'price' or 색 (*sayk*) 'color'. A value indicates a purchaser's specific evaluation of a topic or aspect. The main characteristic of ETs compared with traditional ABSA opinion triples is that values include not only single words and MWEs with a sentiment polarity, but also non-sentiment expressions representing purchasers' evaluations about each aspect. Examples of non-sentiment expressions include 길어요 (*kileyo*) 'it is long' and 검정색이에요 (*kemcengsaykieyyo*) 'it is black'. The following sections describe the individual characteristics of each element of ETs in the clothing domain.

### 3.1 Topics

Topics involve various types of entity names that purchasers evaluate. Table 1 displays the topic classification of the clothing domain. There are five categories. The (CLO_TY) type is the most frequent one, including common nouns of clothing such as 원피스 (*wenphisu*) 'onepiece' and 자켓 (*cakheys*) 'jacket'. The other types are brand names (CLO_BR) such as 나이키 (*naikhi*) 'Nike', online shopping mall names (CLO_SH) such as 지그재그 (*cikucayku*) ‘Zigzag’, store names (CLO_ST) such as 아뜨랑스 (*attulangsu*) ‘Attrangs’, and product parts (CLO_PA) such as 단추 (*tanchwu*) ‘button’.

| Subcategory | Category | Examples |
|---|---|---|
| CLO_TY | 상품타입<br>*Sangphwumthaip*<br>(Cloth Type) | 원피스, 자켓 등<br>*wenphisu, cakheys*<br>(onepiece, jacket etc.) |
| CLO_BR | 브랜드명<br>*Pulayntumyeng*<br>(Brand Name) | 나이키, 폴로 등<br>*naikhi, phollo*<br>(Nike,Polo etc.) |
| CLO_ST | 스토어명<br>*Suthoemyeng*<br>(Store Name) | 아뜨랑스, 리린 등<br>*attulangsu, lilin*<br>(Attrangs, Leelin etc.) |
| CLO_SH | 쇼핑몰명<br>*Syophingmolmyeng*<br>(Shopping Mall Name) | 지그재그, 서울스토어 등<br>*cikucayku, sewulsuthoe*<br>(Zigzag, Seoulstore etc.) |
| CLO_PA | 상품일부<br>*Sangphwumilpwu*<br>(Cloth Part) | 단추, 주머니 등<br>*tanchwu, cwumeni*<br>(button, pocket etc.) |

Table 1: Topics in the clothing domain

### 3.2 Aspect-Value Pairs

An advantage of the ET concept is that aspect and value are considered as paired information, including when the value term occurs without the aspect term. For example, if the predicate 길다 (*kilta*) ‘long’ appears in the evaluation sentence, this word implicitly includes the aspect 길이 (*kili*) ‘length’, which can be explicitly realized within the sentence or not, which is more frequent. Therefore, the aspect-value pair is set to save the implicit aspect with a predicate denoting value if no specific aspect is realized in a sentence.

In this study, aspect-value pairs are classified in three types according to their semantic characteristics: Unary, Binary, and Multiple.

First, Unary includes the aspects with values realized in existential predicates. We divide the Unary type into two subtypes: intrinsic properties of the product, such as elasticity and water resistance, and situational properties, such as external properties.

Second, Binary is the type for evaluation predicates with binary value, such as 길다 (*kilta*) ‘it is long’ and 짧다 (*ccalpta*) ‘it is short’. Most of such measurement adjectives belong to the Binary type.

Third, Multiple is for aspects with more than two values, such as 빨강 (*ppalkang*) ‘red’, 검정 (*kemceng*) ‘black’, and 노랑 (*nolang*) ‘yellow’. The examples in Table 2 show that sentiment and non-sentiment evaluation for an aspect are considered separately.

| Type | Aspect | {ASPECT-VALUE} | Example |
|---|---|---|---|
| UNARY | 방수성<br>*pangswuseng*<br>(Waterproof) | WATERPROOF-GOOD/BAD | 방수성 좋은 소재<br>*pangswuseng coun socay*<br>(It's waterproof material) |
| BINARY | 길이<br>*Kili*<br>(Length) | LENGTH-LONG/SHORT | 기장이 긴<br>*kicangi kin*<br>(The length is long) |
| | | LENGTH-GOOD/BAD | 길이가 적당해요<br>*kilika cektanghayyo*<br>(The length is appropriate) |
| MULTIPLE | 디자인<br>*ticain*<br>(Design) | DESIGN-TYPE | 브이넥<br>*uineyk*<br>(V-neck) |
| | | DESIGN-GOOD/BAD | 디자인이 예쁜<br>*ticaini yeyppun*<br>(The design is pretty) |

Table 2: Example of {ASPECT-VALUE}

## 4 Resource construction

We built linguistic resources using the SSP method, which annotates various language patterns in data with the aid of LGGs. We utilized lexical databases (Deco-Dom) to specify the language patterns in LGGs. The main advantage of the SSP methodology is that it produces training data efficiently in terms of time and cost as compared to crowdsourcing, which has been widely used to create large-scale training data. The following sections provide further details.

### 4.1 Deco-Dom

The Deco-Dom dictionary covers the vocabulary of various domains because users can directly configure it according to the intended purpose or domain. In addition, because it can be configured in a format compatible with the ‘Dictionnaire Électronique du Coréen’ (DECO) Korean lexical database (Nam, 2018), it enables morphological analysis.

| Element category | POS | Examples | Count |
|---|---|---|---|
| TOPIC | noun | 자켓, 주머니 (*cakheys, cwumeni*)<br>(jacket, pocket etc.) | 1844 |
| ASPECT | noun | 길이, 사이즈 (*kili, saicu*)<br>(length, size etc.) | 116 |
| VALUE | adjective | 크다, 길다 (khuta, kilta)<br>(large, long etc.) | 88 |
| | verb | 덮다, 맞다 (*Tephta, macta*)<br>(cover, fit etc.) | 21 |

Table 3: Statistics of the Deco-Dom Dictionary

The DECO lexical database has been created through linguistic studies (Nam, 1996), bearing in mind the following methodological safeguards (Gross, 1989):

- grounding decisions on systematic inventories of lexical entries, not on sporadic observations;
- using readable, updatable data formats from the beginning;
- defining modes of inflection (i.e. the set of morphological changes to a lemma when generating inflected forms) explicitly and independently of one another;
- assigning each entry a code for the applicable mode of inflection.

Korean being an agglutinative language, inflection was modularized in two steps (Berlocher et al., 2006), implemented by Paumier (2003):

- generating morpheme-internal morphological variants, e.g. ㅋ (*kh*), a form of 크 (*khu*) ‘big’, with the method of Wehrli (1985);
- appending suffixes, for instance ㅋ (*kh*) ‘big’ can be followed by the past suffix -었 (-*eoss*), giving 컸 (*kheeoss*) ‘was big’.

The Deco-Dom is an extension that covers domains by inserting domain-specific tags and entries.

### 4.2 Local Grammar Graph

Local Grammar Graph(LGG) is a formalism to describe linguistic patterns in the format of directed word graphs (Gross, 1997). LGGs allow for annotating sequences denoting aspect-value pairs. For example, LGGs process collocations such as 뒤꿈치까지 내려와요 (*twikkwumchikkaci naylyewayo*) ‘It comes down to the heel’ and 짧지 않아요 (*ccalci anayo*) ‘It is not short’. Figure 2 shows a part of LGGs that cover the LENGTH-LONG pair.

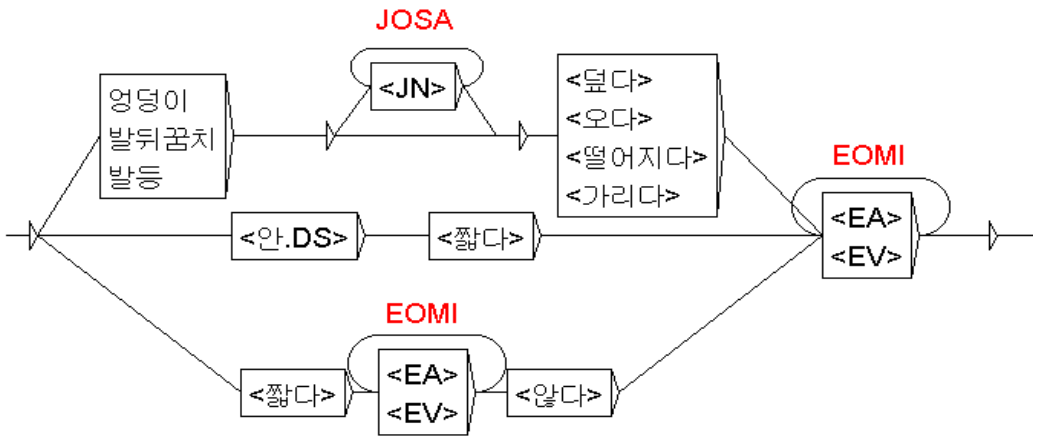

Figure 2: A part of LGGs for LENGTH-LONG

Since an LGG is a directed word graph or finite-state automaton, it is equivalent to a regular expression, but more easily readable and updatable in practice. In its visual form (Fig. 2), it takes advantage of both dimensions: horizontal for sequences of words or phrases, and vertical to enumerate alternatives; whereas a regular expression, in its visual form, linearizes all operations on a single dimension (a formula). Thus, LGGs can describe more complex sets of expressions more clearly.

An LGG can invoke others as subgraphs. This feature allows for managing complexity. As a matter of fact, the above graph is indirectly called by the main graph named ASPECT-VALUE as shown in Figure 3.

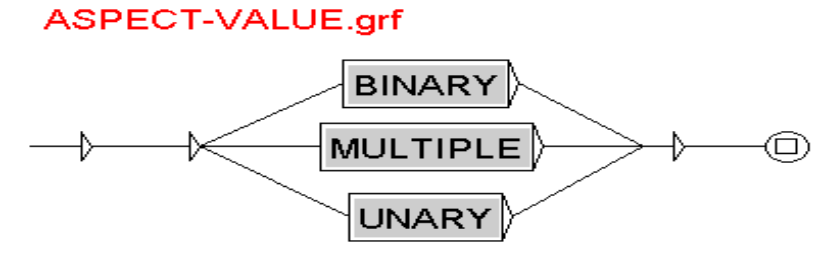

Figure 3: The main LGG for Aspect-Value pairs

Invocations can be recursive, giving LGGs the expressive power of context-free grammars.

LGGs can specify words or word elements either literally, as <짧다> (*ccalpta*) in Fig. 2, or through grammatical symbols, as <JN> which refers to a category of suffix sequences. When the LGGs are used as a query to locate or annotate expressions in texts, the search engine resolves the grammatical symbols on the basis of a lexical database.

The LGG in Figure 3 calls three sets of subgraphs: the subgraph set UNARY which represents 329,153,486 expressions, BINARY with 268,435,457, and the MULTIPLE with 187,856,271 expressions, which allows us to recognize and annotate more than 780 million patterns of aspect-value pairs.

| Aspect-Value pair type | # of patterns |
|---|---|
| UNARY | 329,153,486 |
| BINARY | 268,435,457 |
| MULTIPLE | 187,856,271 |

Table 4: Number of patterns for 3 types of Aspect-Value pairs

Whe the patterns in Table 4 are detected in a given corpus, the LGGs allow the annotation of the related information through the Unitex platform that compiles LGGs into FSTs for efficient application.

### 4.3 SSP-based generation of annotated text

The process defined by the SSP(Semi-automatic Symbolic Propagation) methodology mainly consists of two phases [1] : **Human-driven construction** of language resources such as Deco-Dom and LGGs and **Automatic generation** of annotated datasets based on the resources built in the first phase (Nam 2021b).

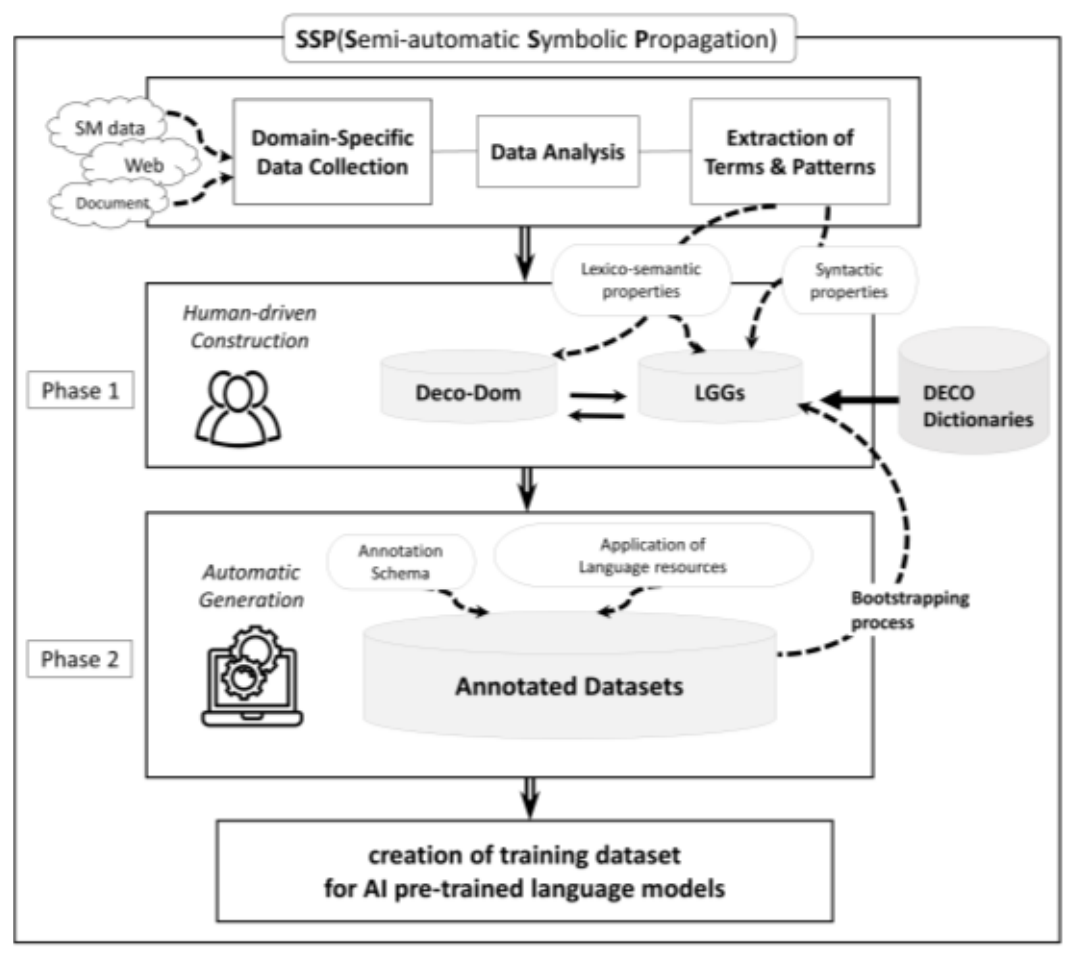

Figure 4: Overview of the SSP methodology

We selected a specific clothing website [2] to collect review data and construct annotated texts. We applied a Deco-LGG resource to approximately 200,000 (199,913) reviews and performed the annotation of the topics and aspect-value pairs through the SSP methodology. Through this processing, we generated evaluation-annotated datasets for fine-grained ABSA. Figure 5 shows examples of EVAD.

<FABRIC-GOOD>재질 괜찮고</FABRIC-GOOD> <SIZE-GOOD>핏이 좋아요</SIZE-GOOD>.

<COLOR-BLACK>블랙으로 샀는데</COLOR-BLACK> <SIZE-GOOD>너무 잘 맞아요</SIZE-GOOD>.

<PRICE-GOOD>저렴하게 잘 샀어요</PRICE-GOOD>.

<ENT=CLO_TY>티셔츠</ENT>는 <LENGTH-SHORT> 기장이 약간 짧네요</LENGTH-SHORT>.

Figure 5: Examples of annotated text sample

ETs in the annotated text EVAD contain detailed information about consumer products. The annotated dataset can be used for a shopping recommendation system which is likely to help consumers make efficient purchase decisions.

## 5 Evaluation of the EVAD dataset

To evaluate the performance of the EVAD dataset, and indirectly of the SSP method used to generate it, we experimented about aspect-value pairs as they are the core of ET. There are frequent aspect-value pairs appearing in the review corpus, such as SIZE-LARGE, FABRIC-THIN, or PRICE-GOOD. For these categories, we evaluated the annotation of EVAD. We used 1,000 sentences extracted from the clothing reviews, manually annotated them with the correct tags, and evaluated the automatically annotated EVAD sentences by comparing them with the test set. Table 5 displays the results.

| | Recall | Precision | F1-Score |
|---|---|---|---|
| EVAD | 0.8804 | 0.8739 | 0.8772 |

Table 5: Evaluation of EVAD

The results in Table 5 show that the EVAD datatset, automatically generated through the SSP methodology, reach a F1-Score of 87%.

## 6 Experiment

We used the KoBERT[3] and KcBERT[4] models in an experiment of training them on EVAD to recognize aspect-value pairs. These models are suitable for this task since they analyze sequences in syllable units. The pre-trained language models KoBERT and KcBERT were used to train the model. Then, the performance of the models was evaluated against the manually annotated test set (1,000 sentences) of Section 5. Table 6 lists the overall performances of the pre-trained language models (PLM).

[1] http://linito.kr/
[2] https://wusinsa.musinsa.com
[3] https://github.com/SKTBrain/KoBERT
[4] https://github.com/Beomi/KcBERT

| | Recall | Precision | F1-Score |
|---|---|---|---|
| KoBERT | 0.8859 | 0.8886 | 0.8873 |
| KcBERT | **0.8990** | **0.9063** | **0.9026** |

Table 6: Evaluation of the PLMs

The PLMs show F1-Scores of 88% and 90%, respectively. The reason why KcBERT performance is slightly better than KoBERT is that its pre-trained data comprises informal texts such as news comments.

## 7 Conclusion

This study proposes an application of the evaluation triple (ET) concept in a clothing domain, and the construction of a linguistic resource, EVAD. The ET consists of three components: topic, aspect, and value. The first two are components of traditional ABSA, whereas the value is introduced to extract detailed information about aspects. We classify 79 aspect-value pairs described from the clothing domain into three types: Unary, Binary, and Multiple.

EVAD is generated with the SSP method and shows a robust F1-score performance of 90%. SSP-based linguistic resources using an ET frame can be applied to various domains. We expect that the concept and method implemented in this study will be a key asset for research on constructing sophisticated annotated training data for deep learning language models.

## Acknowledgments

This work was partially supported by LINITO Ltd. (http://linito.kr) and DICORA Research Center (http://dicora.kr) in Hankuk University of Foreign Studies. We thank the anonymous reviewers for their helpful comments.